# Robust cDNA microarray image segmentation and analysis technique based on Hough circle transform

R. M. Farouk[1] and M. A. SayedElahl[2,*]

[1]Department of Mathematics, Faculty of Science, Zagazig University, Egypt. E-mail: rmfarouk1@yahoo.com.

[2]Department of Mathematics, Faculty of Science, Banha University, Egypt. E-mail: ma0777@yahoo.com.

**ABSTRACT**

One of the most challenging tasks in microarray image analysis is spot segmentation. A solution to this problem is to provide an algorithm than can be used to find any spot within the microarray image. Circular Hough Transformation (CHT) is a powerful feature extraction technique used in image analysis, computer vision, and digital image processing. CHT algorithm is applied on the cDNA microarray images to develop the accuracy and the efficiency of the spots localization, addressing and segmentation process. The purpose of the applied technique is to find imperfect instances of spots within a certain class of circles by applying a voting procedure on the cDNA microarray images for spots localization, addressing and characterizing the pixels of each spot into foreground pixels and background simultaneously. Intensive experiments on the University of North Carolina (UNC) microarray database indicate that the proposed method is superior to the K-means method and the Support vector machine (SVM).

## INTRODUCTION

Microarray technology allows simultaneous measurement of thousands of genes in a single experiment. This provides a useful tool for evaluating the expression of genes and extraction of the characterization and chromosomal structural information about these genes. Microarrays are arrays of glass microscope slides, in which thousands of discrete DNA sequences are printed by a robotic array, thus, forming circular spots of known diameter. Each spot in the microarray image contains the hybridization level of a single gene [1]. Wherever, the amount of the fluorescence hybridization is affected by things that happen during the manufacturing of cDNA microarray images [2], the efficiency of the experimental preparation of the microarray images directly affects the precision of the microarray data analysis [3].

Microarray images processing always pass through three steps: (i) gridding to detect the position of the spot center of the image and identifies their coordinates, (ii) segmentation, which segments, each microarray spot into foreground and background pixels, and (iii) intensity extraction to calculate the foreground fluorescence intensity and background intensities [4].

One important task involved in the analysis of cDNA microarray images is the spots, addressing and gridding, can be divided into three main categories: (i) manual, (ii) semi-automated, and (iii) automated. Many papers have been published presenting different techniques of addressing [5]. Most of these techniques based on the calculation of vertical and horizontal image intensity profile, as presented in the following papers [6, 7, 8, 9].

The other important task in the analysis of cDNA microarray images is a microarray image segmentation process, which characterizes the pixels into foreground pixels and background. Since it considerably affects the precision of micro array data, the segmentation has been a most important and challenging one. The microarray image segmentation techniques can be categorized into four categories (i) Fixed and adaptive circle, considers the spots with circle shape [10], which is used in ScanAlyze and GenePix, (ii) Histogram-based method, it uses a circle target mask to cover all the foreground pixels, and computes a threshold using the Mann-Whitney test [11, 12], (iii) Adaptive shape method, performs image segmentation based on spatial similarity among pixels [13, 14], (iv) Clustering method, as a most common technique, has the advantage that they are not restricted to a particular shape and size for the spots [15]. Since segmentation is used for dividing the image into the regions of foreground and background, the number of cluster centers k is set to two. As the initial cluster centers, the pixels with minimum and maximum intensities are selected. All data points are then assigned to the nearest cluster centers according to a distance measure (e.g., Euclidean distance). Thereafter, new cluster centers are set to the mean of the pixel values in each cluster. Finally, the algorithm is iteratively repeated until the cluster centers stay unaltered [16, 17]. Kernel density estimation KDE can be applied to find their estimated densities after using Gaussian mixture model to found the foreground and background. Then, a cut-off point for segmenting a spot into two clusters is determined by the equality of two estimated densities. The details are given in Algorithm 6.3 [18].

In the present paper a unique cDNA microarray image segmentation framework is introduced, the proposed method based on the CHT algorithm for spots localization, addressing and segmentation simultaneously. The rest of the paper is organized as follows. In the second section, the proposed techniques are explained in detail. Experimental results and the comparisons with the segmentation results of K-means and SVM are discussed in the third section. Finally, the conclusion is analyzed in the fourth section.

## METHODS

The proposed method based on applying CHT algorithm on cDNA microarray for spots localization, addressing and segmentation simultaneously. The input cDNA microarray image comes across several steps before the proposed technique is performed as shown in the flowchart at fig. 1. It typically involves the following steps: (i) Read the cDNA microarray image and crop a spot region of 21 rows x 24 columns. (ii) Smoothing operator (Median) is applied in order to extract gradient information of the cropped image to identify the location of all spots on the microarray image. (iii) Canny edge detection technique [21] is applied in order to extract the edge of each spot in the cropped image. (iv) CHT algorithm [19] is applied in order to get a circle around each spot in the cropped image. (v) Each spot is extracted as individual image, compare the segmentation result of CHT with other older methods and measure the *gene expression* level.

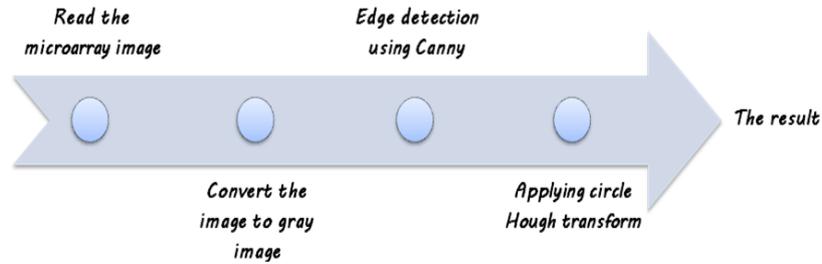

**Fig. 1.** The flowchart of the proposed technique.

**Microarray images database**

Real microarray images from universal microarray images database UNC [20] are employed to evaluate the proposed technique. Pixel by pixel information has been extracted using the annotation file of the UNC images, simulating the fixed circle segmentation. A binary map is generated for the each spot, characterizing the pixels inside the circle as signal pixels and the pixel outside of the circle as background as shown in Fig.3. This binary map is used to evaluate the efficiency of the proposed technique. The properties of the images, we used are shown in table 1.

**Table 1.** The properties of the cropped microarray images

| Experiment ID | Number of spots in Cropped image | Cropped image Size | Number of spots in Cropped image | Spot diameter |
|---|---|---|---|---|
| 42117 | 16915 | 21 x 24 | 504 spots | 50 : 200 |
| 42119 | 16915 | 21 x 24 | 504 spots | 80 : 220 |
| 43431 | 16915 | 21 x 24 | 504 spots | 80 : 220 |
| 41507 | 16128 | 21 x 24 | 504 spots | 60 : 210 |
| 40146 | 16128 | 21 x 24 | 504 spots | 60 : 210 |
| 40141 | 16884 | 21 x 24 | 504 spots | 70 : 200 |

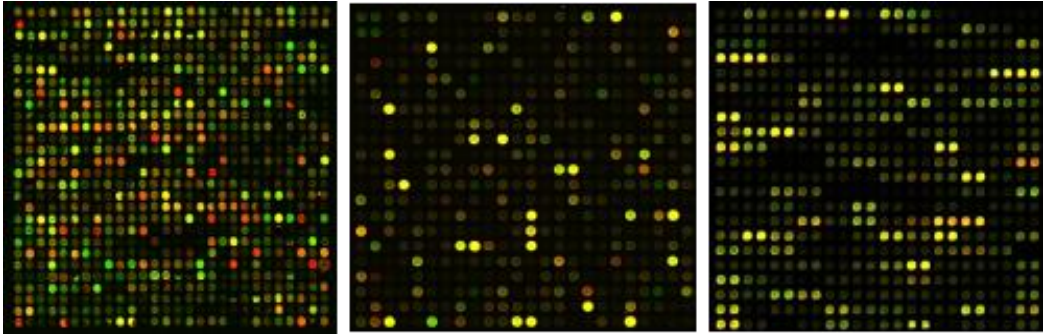

**Fig. 2.** examples of cDNA microarray image.

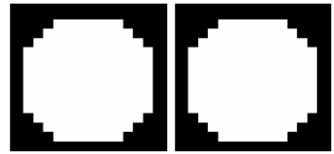

**Fig. 3.** Spots binary map.

**cDNA Microarray Image Filtering.**

All smoothing techniques are effective at removing noise, but adversely affect edges. When reducing the noise, it is important to preserve the edges. Median filtering is a nonlinear image smoothing technique, which can preserve image details well while eliminating noise. Other reasons why we choose the median filter are that it is simple and its calculation complexity is relatively low.

The main idea of the median filter is to run through the signal entry by entry, replacing each entry with the median of neighboring entries. The pattern of neighbors is called the "window", which slides, entry by entry, over the entire signal.

The median is calculated by first sorting all the pixel values from the window in numerical order, and then replacing the pixel being considered with the middle (median) pixel value. Supposing $f(x, y)$ is the gray level of $(x,y)$, then the output of the median filter is described as follows :

$H = Med_W\{f(x,y)\}$

Where W is the filtering window.

Example of Median filter

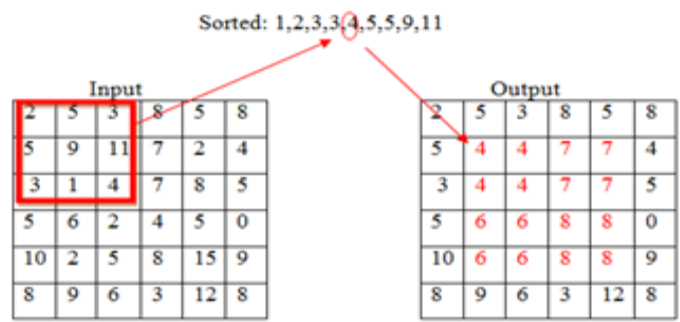

Median filter with window size 3 × 3 is applied on the gray scale microarray image (Fig.4) in order to extract the gradient information of the cropped image and identify the location of all spots on the microarray image, as in Fig. 5.

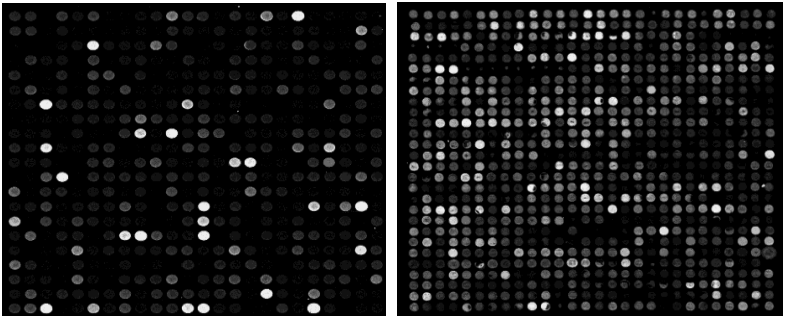

**Fig.4.** cDNA microarray gray scale image.

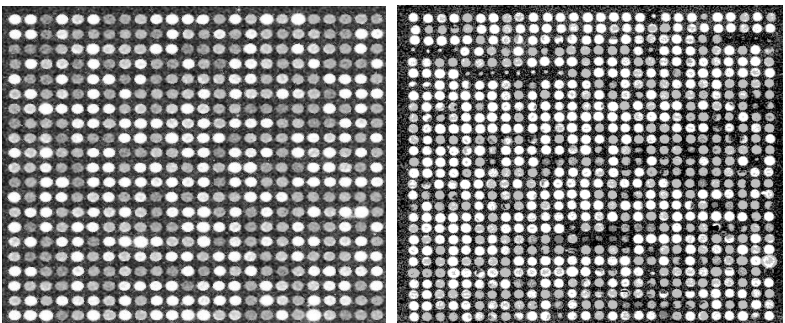

**Fig.5:** cDNA microarray filtered image.

## Edge Detection

Canny edge detection technique [21] is applied in the filtered image (Fig.5) in order to extract the edge of each spot in the image as in Fig.6.

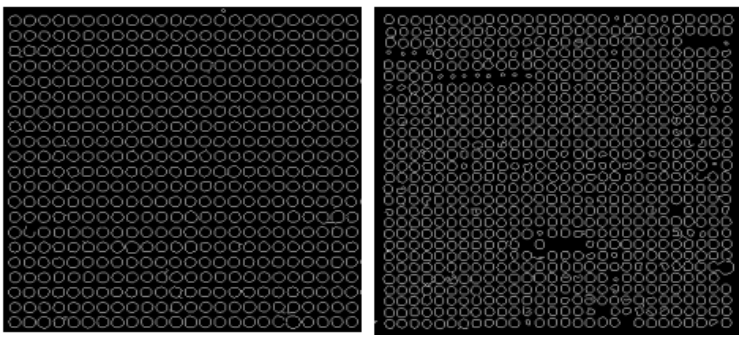

**Fig.6.** Canny edge detection results.

## Circular Hough Transformation (CHT).

Hough transform is recognized as a robust curve detection technique. This method can detect objects, even noise existence. Duda and Hart produced the first sketch of Circular Hough transformation in 1972 [25]. CHT is one of the modified versions of Hough transformation; it aims to find the circular patterns within an image. The main idea of CHT is to transform a set of feature points in the image space into a set of accumulated votes in a parameter space. Then, for each feature point, votes are accumulated in an accumulator array of all parameter combinations. The array elements that contain the highest number of votes indicate the presence of the shape. The circle pattern is described by equation 1. There are many modifications on the CHT either to increase the detection rate or reduce its computational complexity [26,27,28].

**The main algorithm of CHT:**

Consider a point $(x_i, y_i)$ in the image. The general equation of a circle is:

$$(x - u)^2 + (y - v)^2 - r^2 = 0 \quad (1)$$

Where u and v are the coordinates of the center and r is the radius of the circle.

If the gradient angle of the edges is available, then this provides a constraint that reduces the number of degrees of freedom and hence the required size of the parameter space. The direction of the vector from the center of the circle to each edge point is determined by the gradient angle, leaving the value of the radius as the only unknown parameter. Thus, the parametric equations of a circle in polar coordinates are:

$$x = u + r \cos\theta. \quad (2)$$

and

$$y = v + r \sin\theta. \quad (3)$$

Solving for the parameters of the circle we obtain the equations

$$u = x - r \cos\theta. \quad (4)$$

and

$$v = y - r \sin\theta. \quad (5)$$

Now, given the gradient angle Q at an edge point (x,y), we can compute $\cos\theta$ and $\sin\theta$. Note that these quantities may already be available as a by-product of edge detection. We can eliminate the radius from the pair of equations above to yield

$$v = u \tan\theta - x \tan\theta + y. \quad (6)$$

**Circle fitting:**
- First quantize the parameter space for the parameters u and v.
- Then assign the accumulator array A(u,v) to zero.
- Then compute the gradient magnitude G(x,y) and angle θ(x,y).
- For each edge point in G(x,y), increment all points in the accumulator array A(u,v) along the line equation 6 .
- The circles centers on the image correspond to the Local maxima in the accumulator array.

In this work, the CHT is used with varying radius from 6 to 10 pixels to detect the spots in cDNA microarray image. Fig.7. Shows a random picked microarray image after applying CHT algorithm, in which each spot has a green circle around it, after that, each spot can be extracted as individual image as shown in fig. 8.

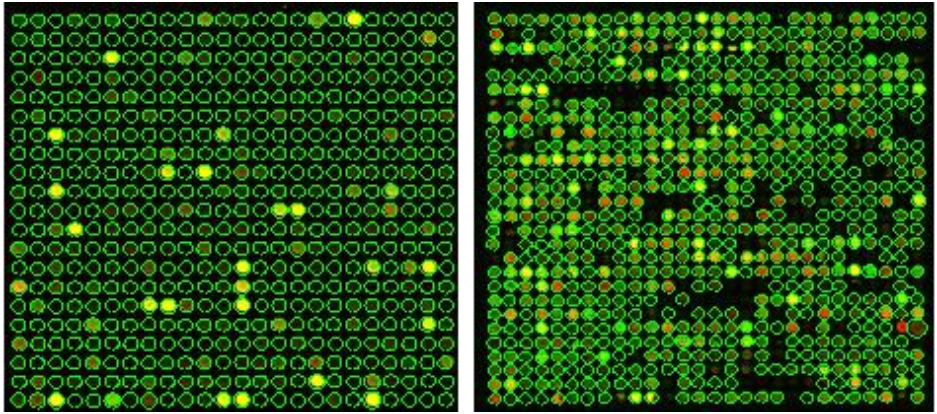

**Fig.7.** The result of applying CHT on the cDNA microarray image.

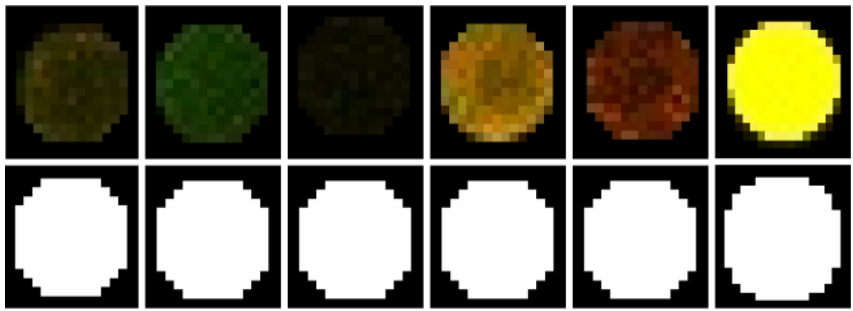

**Fig.8.** Example of the extracted spots and their corresponding binary map.

## EXPERIMENTAL RESULTS

Matlab 2010 [29] is used to create and apply the proposed technique and ran them on the Intel-based Pc with Windows 7 os. The microarray images used in the experiments described in table 1 and all the blocks were stored in TIFF files.

For most of the spots, the background is separated from the foreground perfectly. The result of the segmentation edge is close to the real spot as shown in Fig.8. The randomly chosen original image and its corresponding segmentation result from the UNC database are shown in Fig.9.

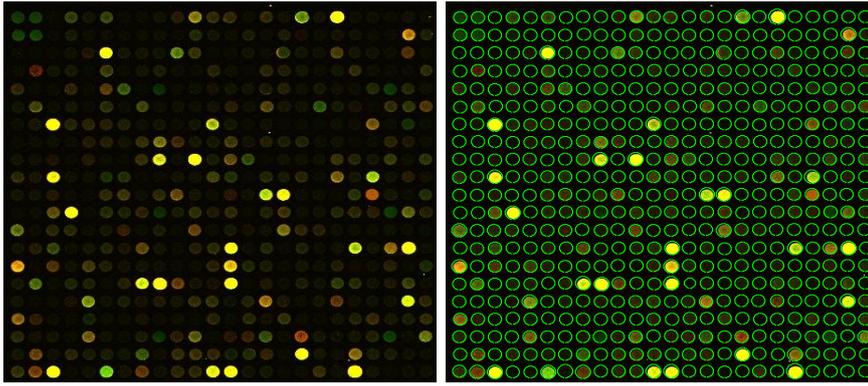

**Fig.9.** Randomly chosen original images in the left side and its corresponding segmentation result on the right side from the UNC database

## COMPARISON OF SPOT SEGMENTATION

cDNA microarray spots are extracted as individual images, as shown in Fig.8. The pixels inside the circle are clustered to signal and outside the circle as background, then compare the segmentation result of the proposed technique with other techniques like K-mean [16] and Support vector machine (SVM) [23].

## K-MEANS CLUSTERING

K-means clustering microarray image segmentation (KMIS) [16]. Attempt to cluster the pixels of an image into two groups (by applying parameter k = 2), one for the foreground, and the other for background. Thus, in KMIS, the feature vector is reduced to a single variable in the Euclidean one-dimensional space. The first step of KMIS consists of initializing the class label for each pixel and calculating the mean for each cluster. Let $x_{min}$ and $x_{max}$ be the minimum and maximum values for the intensities in the spot. If $|x_i - x_{min}| > |x_i - x_{max}|$, $x_i$ is assigned to foreground, or equivalently the label of pixel $x_i$ is set to '1'. Otherwise, $x_i$ belongs to the background, thus $x_i$ is labeled '2'.

## SUPPORT VECTOR MACHINES

Support Vector Machines (SVM) [23] composes a powerful learning system which simultaneously minimizes the empirical classification error and maximizes the geometric margin. Suppose a set of training vectors belonging to two separate classes, { $(y_1, c_1), (y_2, c_2), \ldots, (y_k, c_k)$ }, where each $y_i$ is the feature vector of each pixel of the image. $c_i$ is either 0 (background) or 1 (signal), indicating the class to which the point $y_i$ belongs. The concept is to find a hyperplane which separates the data and can be written as: wx-b=0, where vector w is the perpendicular vector to the hyperplane and the parameter b determines the offset of the hyperplane from the origin along the vector w. In our case SVMs produce the maximal-margin hyperplane which divides the background from the signal. Margin refers to the distance between the hyperplane and the nearest data point in each class.

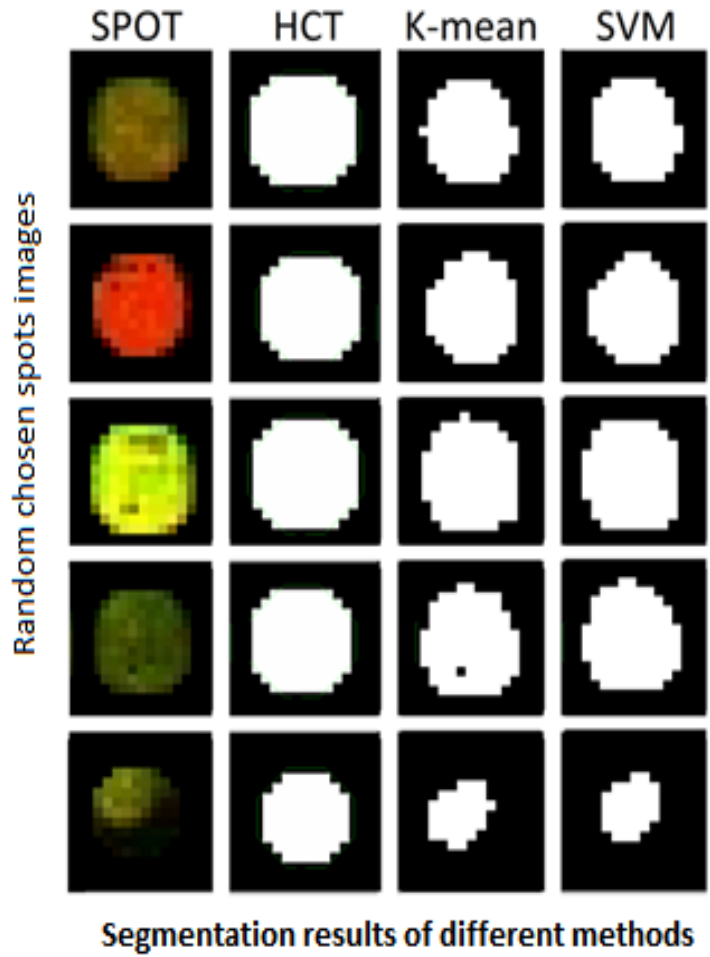

**Fig.10**.Comparison the results of three segmentation methods CHT, K-means, and SVM .

From Fig.10, one can observe that the proposed method optimally segmented most of the microarray spots. When taking the k-means or SVM and our method into account, we can find obviously that the performance of our method is far better than the k-means or SVM, that is, the proposed method is better than the k- means method. The proposed method is closer to the real spot than the other one.

## COMPARISON OF GENE EXPRESSION

In this part, we compare the results of the proposed method with the results of k-means method and SVM using Gene expression level [30].

For the two channels, green and red, the expression values are defined, by Green intensity and Red intensity as the following:

For the green channel

Green intensity = (mean value of the foreground – mean value of the background)     (7)

For the red channel

Red intensity = (mean value of the foreground – mean value of the background)     (8)

Gene expression level = $\log_2$ (Red intensity / Green intensity)     (9)

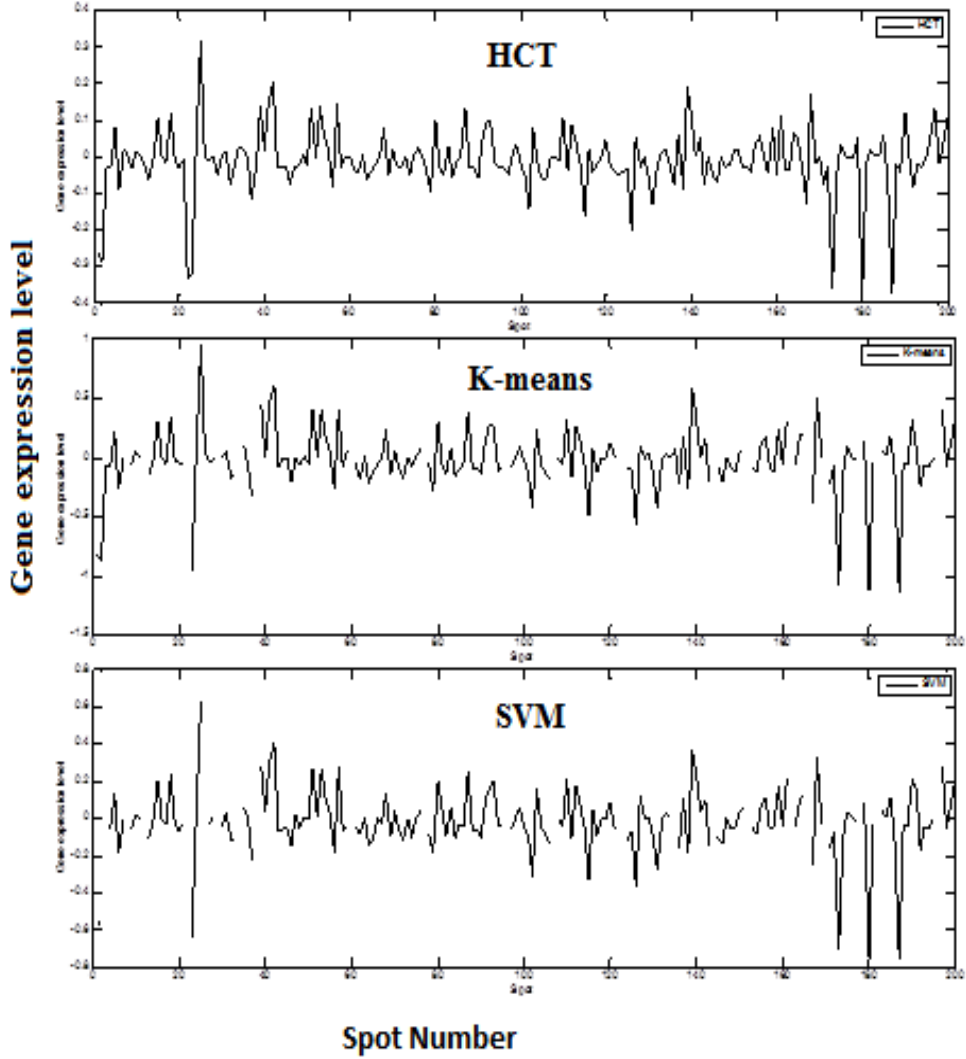

**Fig.11**. Gene expression level after using the different segmentation methods (CHT – K-means – SVM)

From Fig.11 we can notice that, the value of gene expression level of CHT variable from -0.4795 to 0.3177 for SVM varying from -0.9476 to 0.6354, and for K-means varies from -1 to 0.9531, this shows that the CHT has the best result comparing with the other methods.

## COMPARISON OF QUALITY INDEX

The Quality index (Q-index) [24] is used to evaluate the result of segmentation of the proposed method as in Fig.12. Q-index is the mean value of The combined quality index ($q_{com2}$) of the two channels.

$$q_{com2} = \left( q_{sig\text{-}noise} * q_{bkg1} * q_{bkg2} \right)^{1/3} \qquad (10)$$

Where $q_{sig\text{-}noise}$ is the signal-to-noise ratio, $q_{bkg1}$ is the local background variability, and $q_{bkg2}$ is the level of the local background.

$$q_{sig\text{-}noise} = \frac{F_{mean}}{F_{mean} + B_{mean}} \qquad (11)$$

$$q_{bkg1} = \frac{1}{\max\left[\frac{BSD}{B_{mean}}\right]} * \frac{1}{\frac{BSD}{B_{mean}}} \qquad (12)$$

$$q_{bkg2} = \frac{1}{\max\left[\frac{bgk0}{bgk0+Bmean}\right]} * \frac{1}{\frac{bgk0}{bgk0+Bmean}} \qquad (13)$$

Where $F_{mean}$ and $B_{mean}$ are the mean value of the signal and the local background respectively, BSD is the standard deviation of the local background, and bgk0 is the global average of the background.

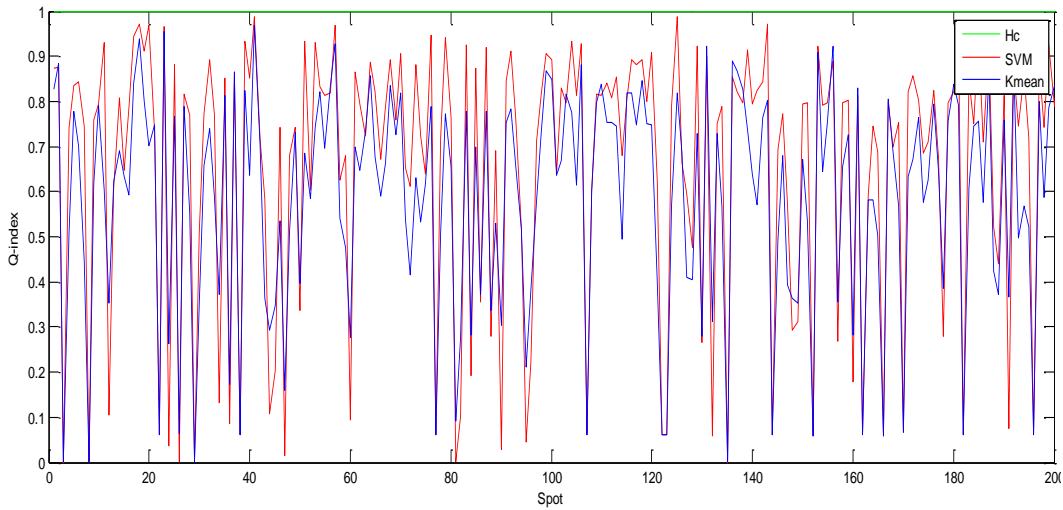

**Fig.12.** The Q-index of the segmentation methods (CHT – SVM – K-means) using fixed circle segmentation(simulated using image information in the UNC database )

Q-index results show that the CHT has the best result comparing with the other methods, it has the same result of circle segmentation using the database accumulation file.

## CONCLUSION

This paper presents a new method for cDNA microarray image analysis using circular Hough transforms. This new method reduces the processing time of spots, addressing and localization and increases the efficiency of spot segmentation. The numerical and visual results show that, the proposed technique is an effective in spot addressing, localization and segmentation.

## REFERENCES

[1] Fenstermacher D. Introduction to Bioinformatics, Journal of the American Society for Information Science and Technology, 56 (5), 440-446, (2005).

[2] Abdul Ahad H. Biometrics-The Human Password, JITPS, 1 (1), 29–42, (2010).

[3] Chee M., Yang R. and Hubbell E., Accessing genetic information with high-density DNA arrays, Science, 610–614, (1996).

[4] Y. H. Yang, M. M. Buckley, S. Dudoit, and T. Speed, "Comparison of methods for image analysis on cDNA microarray data," J. Compo Graph. Stat., pp.1 08-136, (2002).

[5] Ye R., Wang T., Bedzyk L. , Croker K., Applications of DNA microarrays in microbial systems, Journal of Microbiological Methods, 47, 257–272, (2001).

[6] N. Giannakeas, F. Kalatzis, M. G. Tsipouras, and D. I. Fotiadis, Spot addresses for microarray images structured in hexagonal grids, Computers methods and programs in biomedicine, 106 (1), 1–13, (2012).

[7] J. D. and Thomas T.,Automatic Gridding of DNA Microarray Images using Optimum Subimage, International Journal of Recent Trends in Engineering ,1 (4), (2009).

[8] Rueda L. and Rezaeian I. A fully automatic gridding method for cDNA microarray images, BMC Bioinformatics, 12-113, (2011).

[9] N. Giannakeas and D. I. Fotiadis, An automated method for gridding and clustering-based segmentation of cDNA microarray images, Computerized Medical Imaging and Graphics 33, 40–49, (2009).

[10] Karim R., Mahmud S., A review of image analysis techniques for gene spot identification in cDNA Microarray images, International Conference of Next Generation Information Technology, (2011).

[11] A. Ahmed, M. Vials, NG. Iyer, C. Caldas, JD. Brenton, Microarray segmentation methods significantly influence data processing, Nucleic Acids Res. 32, 50-58, (2004).

[12] Y. Chen, E.R. Dougherty, and M.L. Bittner, "Ratio-Based Decisions and the Quantitative Analysis of cDNA Microarray Images, Journal Of Biomedical Optics vol.2(4), pp.364–374, (1997).

[13] M.J. Buckley, Spot User's Guide, CSIRO Mathematical and Information Sciences, Sydney, Australia, (2000).

[14] K.I. Siddiqui, A. Hero, and M. Siddiqui, Mathematical Morphology applied to Spot Segmentation and Quantification of Gene Microarray Images, Asilomar conference on Signals and Systems, (2002).

[15] D. Bozinov, and J. Rahnenfuhrer, Unsupervised technique for robust target separation and analysis of DNA microarray spots through adaptive pixel clustering, Bioinform,, vol. 18, pp. 747–756, (2002).

[16] E. Ergüt, Y. Yardimci, E. Mumcuoglu, O. Konu, Analysis of microarray images using FCM and K-means clustering algorithm, in Proc IJCI, pp.116-121, 2003.

[17] W. Shuanhu and H. Yan, Microarray Image Processing Based on Clustering and Morphological Analysis, The First Asia-Pacific bioinformatics conference on Bioinformatics -Australia, (19), 111-118, (2003).

[18] T. B. Chen, H. H. S. Lu, and Y. J. LAN, Segmentation of cDNA microarray images by kernel density estimation, 41 (6), 1021–1027, (2008).

[19] S. Pei and J. Horng, Circular arc detection based on Hough transform, Pattern Recognition Letters, Volume 16 (6), 615–625,( June 1995).

[20] University of North Carolina (UNC) microarray database https://genome.unc.edu

[21] M. Accame and G.B. De Natale, Edge detection by point classification of Canny filtered images, Signal Processing, 60, (1), 11–22, (July 1997).

[22] GenePix 4000 A User's Guide, (1999).

[23] N. Giannakeas, P.S. Karvelis and D.I. Fotiadis, A Classification-Based Segmentation of cDNA Microarray Images using Support Vector Machines, 30th Annual International IEEE EMBS Conference, Canada, (2008).

[24] A. Liew, H. Yanand M. Yang, Robust adaptive spot segmentation of DNA micro-array images, Pattern Recognition, 36 (5), 1251–1254, (2003).

[25] R.O. Duda and P.E Hart, Use of the Hough transformation to detect lines and curves in picture. Commun. ACM, 11-15, (1972).

[26] C. Kimme, D. Ballard and J. Sklansky, Finding circles by an array of accumulators. Proc. ACM, 18: 120-122, (1975).

[27] D. Ioannou, W. Duda and F. Laine, Circle recognition through a 2D Hough Transform and radius histogramming. Image and Vision Computing, 17: 15-26, (1999).

[28] Guil, N. and E.L. Zapata, Lower order circle and ellipse hough transform. Pattern Recognition, 30: 1729-1744, (1997).

[29] MathWorks, Image Processing Toolbox 7: User's Guide, (2010).

[30] A. L. Tarca, R. Romero and S. Draghici, Analysis of microarray experiments of gene expression profiling, American Journal of Obstetrics and Gynecology 195, 373–88, (2006).